\documentclass{iaf-iac}
\usepackage[backend=bibtex,giveninits=true,style=nature,sorting=none,citestyle=numeric-comp, maxcitenames = 6]{biblatex}
\AtEveryBibitem{\clearfield{note}}
\AtEveryBibitem{\clearfield{language}} 

\usepackage{balance}
\usepackage{placeins}
\usepackage{cuted}
\usepackage{caption}
\usepackage{acronym}
\usepackage{adjustbox}
\usepackage[dvipsnames]{xcolor}
\usepackage{amsmath}

\usepackage{amssymb}
\usepackage{anyfontsize}
\usepackage[english]{babel}
\usepackage{booktabs}
\usepackage{color}
\usepackage{ctable}
\usepackage{eso-pic}
\usepackage{float}
\usepackage{floatflt}
\usepackage[T1]{fontenc}
\usepackage{graphics}
\usepackage{hyperref}
\usepackage{longtable}
\usepackage{makeidx}
\usepackage{makecell}
\usepackage{mathtools}
\mathtoolsset{showonlyrefs=true}
\usepackage{multirow}
\usepackage{notoccite}
\usepackage{setspace}
\usepackage{subfigure}
\usepackage{url}
\usepackage{wasysym}
\usepackage{wrapfig}
\usepackage{newtxmath}
\usepackage{microtype}

\usepackage{indentfirst}

\renewcommand{\phi}{\varphi}
\renewcommand{\epsilon}{\varepsilon}

\graphicspath{{figures/}}

\acrodef{DFT}{discrete Fourier transform}
\acrodef{NEP}{nuclear electric propulsion}
\acrodef{SEP}{solar electric propulsion}
\acrodef{OTV}{orbital transfer vehicle}

\title{
Integrating Low-Thrust Transfer Dynamics into POMDP Planning for On-Orbit Servicing}
\author{
Susmitha Patnala$^{\textbf{a}}$,
Vincent Mousseau$^{\textbf{b}}$,
Adam Abdin$^{\textbf{c}}$\thanks{}}

\begin{document}

\twocolumn[
\begin{@twocolumnfalse}

\centering
IAC-26,C1,IPB,29,x110475

\maketitle

\begin{flushleft}

$^{\textbf{a}}$
\textit{Université Paris-Saclay, CentraleSupélec,
Department of Industrial Engineering,
Gif-sur-Yvette, France,}
\href{mailto:susmitha.patnala@centralesupelec.fr}
{susmitha.patnala@centralesupelec.fr}
\\

$^{\textbf{b}}$
\textit{Université Paris-Saclay, CentraleSupélec, Department of Mathematics and Informatics,
Gif-sur-Yvette, France,}
\href{mailto:vincent.mousseau@centralesupelec.fr}
{vincent.mousseau@centralesupelec.fr}
\\

$^{\textbf{c}}$
\textit{Université Paris-Saclay, CentraleSupélec,
Department of Industrial Engineering,
Gif-sur-Yvette, France,}
\href{mailto:adam.abdin@centralesupelec.fr}
{adam.abdin@centralesupelec.fr}
\\

$^{*}$ Corresponding author

\end{flushleft}

\begin{abstract}

Recent developments in both high-thrust (HT) and low-thrust (LT) propulsion technologies broaden design options for on-orbit servicing (OOS) missions. However, incorporating LT transfers into OOS mission planning requires accounting for the trade-off between prescribed transfer duration and propellant consumption.  Multi-servicing, multi-transfer OOS mission planning is inherently sequential and requires adaptive planning tools that account for propulsion-dependent transfer durations and mass-dependent propellant consumption. Thus, transfer dynamics must be integrated into mission planning and scheduling to increase flexibility and adaptability.  In addition, the planning is often done with imperfect information about the system's state and stochastic service demands.    
This paper focuses on modeling orbital transfers, particularly LT transfers, within a Partially Observable Markov Decision Process (POMDP) framework for OOS mission planning. To enable offline policy computation, the transfer model must provide explicit state transitions and rewards for all admissible actions. The state space is therefore augmented with the servicing vehicle's remaining propellant to enable consistent mass updates. HT transfers are modeled using a precomputed transfer matrix that maps each transfer to the time of flight and propellant consumed. LT transfers are represented using an analytical GEO phasing model with prescribed transfer-duration alternatives. For each alternative, the
model calculates propellant consumption and maneuver feasibility as functions
of the required phase change and the current servicing-vehicle mass.

The proposed modeling approach is evaluated on simplified OOS planning scenarios using circular phasing with HT and LT propulsion. The resulting POMDPs are solved using the Successive Approximations of the Reachable Space under Optimal Policies (SARSOP) algorithm, and the policies obtained under HT and LT are compared. This comparison provides a quantitative assessment of the feasibility and operational impact of LT options, particularly regarding the potential role of advanced electric propulsion in future servicing architectures.

\end{abstract}

\begin{flushleft}
\end{flushleft}

\end{@twocolumnfalse}
]

\section{Introduction}

On-orbit servicing (OOS) includes activities such as inspection,
maintenance, repair, refueling, repositioning, and life extension of satellites \cite{caramelli2023space}. These capabilities can increase the operational lifetime and flexibility of space systems while reducing the reliance on frequent launches. However, in a multi-target servicing
mission, each servicing decision changes the location and remaining
resources of the servicing vehicle and influences which future service
requests can be addressed \cite{abdin2025strategic}. Consequently, the value of a maneuver depends on the sequence in which target satellites are visited and on the operational state of the constellation when the maneuver is initiated \cite{long2007orbit,verstraete2018,sarton2021framework}.

Propulsion introduces an important fuel-time trade-off into this
sequential planning problem. Chemical high-thrust (HT) propulsion
provides relatively short transfer times but generally consumes more
propellant. Electric low-thrust (LT) propulsion provides substantially
higher specific impulse and can reduce propellant expenditure, but its
lower acceleration results in longer transfer durations. LT transfers can therefore increase the time for which failed satellites
remain unattended and the probability that other satellites degrade
while the servicing vehicle is in transit.

LT trajectory performance also depends on the current mass of the servicing spacecraft. For a given phase change and time of flight, the initial mass affects the duration of thrust, propellant consumption, and the feasibility of the maneuver. Previous studies have investigated
LT transfers for geosynchronous satellite servicing
\cite{han2019}, and the use of high-thrust, low-thrust, and
multimodal vehicles in OOS logistics optimization
\cite{sarton2022orbit}. These studies establish the operational importance of the propulsion trade-off but do not account for imperfect knowledge of satellite health in the sequential decision state.

In practice, the health of a target satellite can only be known through incomplete telemetry, noisy monitoring, or an inspection.
The servicing vehicle must then balance the value of acquiring
additional information against the cost of delaying maintenance
action. Propulsion and partial observability interact directly. During
a long transfer, unobserved targets continue to degrade, existing
health information becomes older, and failed satellites can accumulate
additional downtime. A fuel-efficient transfer may therefore be operationally undesirable when the probability of failure is high, whereas a rapid high-thrust transfer may unnecessarily consume scarce
propellant when service urgency is low.

In previous work, an event-based POMDP framework was developed for
inspection and maintenance planning under partial observability
\cite{patnala2026optimal}. The formulation represented uncertain
satellite health, noisy inspections, stochastic degradation,
servicing-vehicle location, remaining propellant, and
action-dependent event durations. The present work extends this model
by introducing propulsion-dependent HT and LT GEO phasing transfers.

The contributions of this paper are as follows:

\begin{enumerate}
    \item A fixed-duration analytical LT GEO phasing model, with mass-dependent
propellant consumption and maneuver feasibility, is integrated into the
event-based POMDP model.

    \item The transfer duration and propellant consumption returned by the trajectory model are incorporated into the health, fuel, and reward transitions of the POMDP.

    \item HT and LT servicing policies are
    compared for a three-satellite GEO scenario under partial observability.

\end{enumerate}

The remainder of the paper is organized as follows. Section
\ref{sec:pomdp} presents the POMDP model. Section
\ref{sec:transfer} describes the HT and LT transfer models and their
integration into the POMDP. Section \ref{sec:solution} presents the
solution method. Section \ref{case_study} defines the numerical case
study, Section \ref{sec:results} discusses the results, and Section
\ref{sec:conclusion} concludes the paper.

\section{POMDP Model}
\label{sec:pomdp}
The servicing problem is formulated as a semi-Markov event-based POMDP. Decisions are made at event epochs
$e=0,1,2,\ldots$, where an event corresponds to the completion of an
idle, inspection, maintenance, or travel action. The model is defined by

\begin{equation}
\mathcal{P}
=
\left(
\mathcal{S},
\mathcal{A},
\mathcal{O},
\mathcal{T},
\mathcal{Z},
\mathcal{R},
\Delta t,
\gamma
\right),
\label{eq:pomdp_tuple}
\end{equation}

where $\mathcal{T}$, $\mathcal{Z}$, and $\mathcal{R}$ are the
transition, observation, and event-reward functions, respectively;
$\Delta t(s,a)$ is the deterministic duration of an event; and
$\gamma\in(0,1)$ is the per-day discount factor. Thus, the continuation
factor after an event of duration $\Delta t_e$ is
$\gamma^{\Delta t_e}$ rather than a fixed discount applied independently
of the action duration.

\subsection{State and belief}

For a constellation of $\mathcal{M}=\{1,\ldots,m\}$ target satellites, the system state at
event $e$ is

\begin{equation}
s_e
=
\left(
H^e,\ell^e,F_e
\right),
\qquad
H^e
=
\left(
h_1^e,\ldots,h_m^e
\right).
\label{eq:state}
\end{equation}

where $h_{i}^e\in\{0,1\}$ is the health state of satellite $i$.
State 0 represents an operational or as-good-as-new satellite, while
state 1 represents a failed satellite requiring service. The variable
$\ell_e\in\{1,\ldots, m\}$ denotes the satellite with which the
servicing vehicle is currently co-located. The discrete variable
$F_e\in\{0,1,\ldots,N_F\}$ represents the remaining propellant. For a fuel-bin size $\Delta_F$, the corresponding propellant mass is

\begin{equation}
m_{\mathrm{prop}}(F_e)=F_e\Delta_F.
\label{eq:fuel_mass}
\end{equation}

The servicer location and propellant are assumed to be known exactly, whereas the satellite health vector is partially observable. The state
can therefore be separated into the fully observable component

\begin{equation}
X_e=(\ell_e,F_e)
\end{equation}

and the partially observable component

\begin{equation}
Y_e=H_e.
\end{equation}

The agent maintains a belief over the joint health state, 

\begin{equation}
b_e(H)
=
\Pr
\left(
H_e=H
\mid
a_{0:e-1},o_{1:e},X_{0:e}
\right).
\label{eq:belief}
\end{equation}

\subsection{Action space}

At each event epoch $e$, the servicing vehicle selects an action
$a^e\in\mathcal{A}$, where

\begin{equation}
\begin{aligned}
\mathcal{A}
={}&
\{\mathrm{NA},\mathrm{M}\}
\cup
\{I_i\mid i\in\mathcal{M}\}
\\
&\cup
\left\{
\mathrm{Travel}_{i\rightarrow j}^{f}
\;\middle|\;
i\neq j,\;
f\in\{\mathrm{low},\mathrm{med},\mathrm{high}\}
\right\}.
\end{aligned}
\label{eq:action_space}
\end{equation}

The no-action decision $\mathrm{NA}$ advances the system by a fixed idle duration. The action $I_i$ inspects satellite $i$ and generates an observation of its health state. The maintenance action $\mathrm{M}$ restores the satellite co-located with the servicing vehicle to the as-good-as-new state.

The action $\mathrm{Travel}_{i\rightarrow j}^{f}$ transfers the
servicing vehicle from satellite $i$ to satellite $j$. The travel mode
$f$ represents a point on the time--propellant trade-off:
$\mathrm{low}$ denotes a long-duration, fuel-efficient transfer,
$\mathrm{med}$ denotes an intermediate transfer, and $\mathrm{high}$
denotes a short-duration, fuel-intensive transfer.

The same action notation is used for the HT and LT case
studies. In the HT case, the action duration and propellant
consumption are obtained from the maneuver matrix. In the LT case, they are obtained from the LT transfer model.
Thus, the two propulsion architectures produce different POMDP models but retain the same state, action, observation, and reward definitions.

\subsection{Health transition model}

Satellite degradation follows the constant-rate model used in the
numerical experiments of our previous work \cite{patnala2026optimal}. For an
operational satellite, the probability of failure during an event of
duration $\Delta t^e$ is

\begin{equation}
\Pr
\left(
h_i^{e+1}=1
\mid
h_i^e=0,a^e
\right)
=
1-\exp
\left(
-\lambda_{\mathrm{deg}}\Delta t^e
\right).
\label{eq:degradation}
\end{equation}

A failed satellite remains failed unless maintenance is performed.
Maintenance resets the co-located satellite to the as good as new
state at the completion of the maintenance event. All other satellites
continue to degrade during idle, inspection, maintenance, and travel
events.

Equation \eqref{eq:degradation} provides the principal coupling between the propulsion and health models: HT and LT transfers between the same
satellites may produce different failure probabilities because their event durations differ.
\subsection{Observation and belief update}

An inspection of satellite $i$ returns a binary observation
$o_i\in\{0,1\}$. The observation model is

\begin{equation}\label{eq:obs_matrix}
Z_i(o_i\mid h_i,I_i)
=
\begin{cases}
p_{\mathrm{obs}}, & o_i=h_i,\\
1-p_{\mathrm{obs}}, & o_i\neq h_i.
\end{cases}
\end{equation}

Co-located inspections are modeled as perfect by setting
$p_{\mathrm{obs}}=1$ when $i=\ell_e$, while remote inspections use the imperfect observation matrix given in Eq. \eqref{eq:obs_matrix}. 

Following action $a_e$ and observation $o_{e+1}$, the belief is
updated using Bayes' rule:

\begin{equation}
b_{e+1}(s')
=
\eta
Z(o_{e+1}\mid s',a_e)
\sum_{s\in\mathcal{S}}
T(s'\mid s,a_e)b_e(s),
\label{eq:belief_update}
\end{equation}

where $\eta$ is a normalization constant.
\subsection{Reward model}

The reward function follows the formulation used in our previous work
\cite{patnala2026optimal}. It includes the reward obtained from successfully
maintaining a failed satellite and the costs associated with
maintenance, inspection, propellant consumption, depreciation, satellite
downtime, and other duration-dependent penalties.

Since the actions have different durations, the costs that accumulate
during an event are discounted over the event duration. Let $\beta=-\ln(\gamma)$,
where $\gamma$ is the per-unit-time discount factor. The discounted
duration associated with an event of duration $\Delta t^e$ is

\begin{equation}
G(\Delta t^e)
=
\frac{
1-\exp\left(-\beta\Delta t^e\right)
}{
\beta
}.
\label{eq:discounted_event_duration}
\end{equation}

The event reward is defined as

\begin{equation}
\begin{split}
\mathcal{R}(s_e,a^e)
={}&
c_{\mathrm{reward}}
\mathbf{1}
\left\{
a^e=\mathrm{M},
h_{\ell^e}^e=1
\right\}
-
c_{\mathrm{M}}
\mathbf{1}
\left\{
a^e=\mathrm{M}
\right\}
\\
&-
c_{\mathrm{obs}}
\mathbf{1}
\left\{
a^e=I_i
\right\}
-
c_{\mathrm{fuel}}m_p(s_e,a^e)
\\
&-
\left[
c_{\mathrm{d}}
+
c_{\mathrm{down}}
N_{\mathrm{failed}}(s_e)
\right]
G(\Delta t^e),
\end{split}
\label{eq:reward}
\end{equation}

where

\begin{equation}
N_{\mathrm{failed}}(s_e)
=
\sum_{i\in\mathcal{M}}
\mathbf{1}
\left\{
h_i^e=1
\right\}
\label{eq:number_failed}
\end{equation}

is the number of failed satellites at the beginning of the event. 
In the present implementation, the duration-dependent downtime term uses
the number of satellites failed at the beginning of the event. This provides an initial approximation of the total accumulated downtime cost and may underestimate failures that happen during extended events. The failed-satellite exposure discussed in Section~\ref{subsec:solver_results} is instead calculated using the realized health trajectory over the full event duration, and thus serves as a separate rollout diagnostic.

The propellant consumption $m_p(s_e,a^e)$ is returned by the HT or LT
transfer model for travel actions and is zero for non-travel actions.

\section{High- and Low-Thrust Transfer Modeling}
\label{sec:transfer}

The target satellites and servicing vehicle are assumed to occupy a
common circular GEO orbit. Transfer-time alternatives are generated for each ordered satellite pair.
\subsection{High-thrust GEO phasing}

The high-thrust maneuver model follows the circular-orbit phasing
approach developed in our previous work \cite{patnala2026optimal}. The target
satellites and servicing vehicle are assumed to occupy a common
circular GEO and to be separated only by their phase angles. Orbital motion is represented using two-body Keplerian dynamics without perturbations.

High-thrust phasing maneuvers are precomputed for every ordered pair
of target satellites $(i,j)$. The candidate maneuvers are classified
into three travel modes, $f\in\{\mathrm{low},\mathrm{med},\mathrm{high}\}$, and a travel matrix is formed. 
For each ordered satellite pair and travel mode, the transfer matrix provides
 $\Delta t_{ij}^{f}$ (the transfer duration) and
$\Delta v_{ij}^{f}$ (the total velocity increment required) by the
two-impulse phasing maneuver.
When a travel action is evaluated, $\Delta v_{ij}^{f}$ is converted
to propellant consumption using the rocket equation and the current
servicing-vehicle mass. The resulting transfer duration and propellant
consumption are incorporated into the POMDP transition and reward
models. The HT formulation is used as the baseline against which the
LT transfer model is compared.

\subsection{Low-thrust GEO phasing}
The low-thrust transfer model adopts the near-circular analytical rephasing approximation presented by \textcite{martinez2015analytical}. The maneuver consists of three phases:

\begin{enumerate}
    \item an initial tangential-thrust spiral from GEO to an
    intermediate approximately circular orbit;

    \item a coast phase during which the required relative phase is
    accumulated; and

    \item a final tangential-thrust phase that returns the servicer to
    GEO at the desired phase.
\end{enumerate}

The model assumes constant, continuous tangential thrust, constant
specific impulse, a thrust acceleration much smaller than the
gravitational acceleration, and approximately constant spacecraft mass
during one maneuver.

Let $F_T$ be the LT thrust magnitude. The signed tangential
acceleration is given by Eq. \eqref{eq:lt_acceleration}.

\begin{equation}
a_\theta
=
-
\operatorname{sign}(\Delta\theta_{ij})
\frac{F_T}{m_0}.
\label{eq:lt_acceleration}
\end{equation}

Under the spiral approximation, the relative phase dynamics are given by Eq. \eqref{eq:relative_phase_dynamics}.

\begin{equation}
\frac{\mathrm{d}^2\delta\theta}{\mathrm{d}t^2}
=
-\frac{3a_\theta}{r_0}.
\label{eq:relative_phase_dynamics}
\end{equation}

Let the initial and final thrust phases have equal duration $\tau$,
and let the prescribed total transfer duration be $t_f$. Integrating
Eq. \eqref{eq:relative_phase_dynamics} over the three maneuver phases
gives Eq. \eqref{eq:lt_quadratic}.

\begin{equation}
\tau^2
-
t_f\tau
+
\frac{
r_0m_0|\Delta\theta_{ij}|
}{
3F_T
}
=
0.
\label{eq:lt_quadratic}
\end{equation}

The physically relevant solution is

\begin{equation}
\tau
=
\frac{1}{2}
\left[
t_f
-
\sqrt{
t_f^2
-
\frac{
4r_0m_0|\Delta\theta_{ij}|
}{
3F_T
}
}
\right].
\label{eq:lt_tau}
\end{equation}

The maneuver is feasible only if the discriminant is nonnegative:

\begin{equation}
t_f^2
-
\frac{
4r_0m_0|\Delta\theta_{ij}|
}{
3F_T
}
\geq0.
\label{eq:lt_feasibility}
\end{equation}

Equivalently, the initial mass must satisfy

\begin{equation}
m_0
\leq
M^{\mathrm{ub}}
=
\frac{
3F_Tt_f^2
}{
4r_0|\Delta\theta_{ij}|
}.
\label{eq:mass_upper_bound}
\end{equation}

Equation \eqref{eq:mass_upper_bound} shows that short-duration
transfers and large phase changes impose the most restrictive mass
limits.

The total LT propellant consumption is derived as Eq. \eqref{eq:lt_propellant}.

\begin{equation}
m_{p,\mathrm{LT}}
=
2\dot{m}\tau
=
2
\frac{
F_T
}{
g_0I_{sp,\mathrm{LT}}
}
\tau.
\label{eq:lt_propellant}
\end{equation}

The present framework can evaluate this nonlinear expression directly at every discrete fuel state.
\subsection{Integration of transfer dynamics into the POMDP}

The initial servicing-vehicle mass is calculated as

\begin{equation}
m_0(F_e)
=
m_{\mathrm{dry}}
+
m_{\mathrm{payload}}
+
m_{\mathrm{prop}}(F_e).
\label{eq:servicer_mass}
\end{equation}

Remaining propellant is state-dependent and
therefore changes the mass supplied to the trajectory model throughout
a servicing sequence.

For each travel action, the trajectory module:

\begin{enumerate}
    \item calculates $m_0$ using Eq. \eqref{eq:servicer_mass};

    \item obtains the phase angle separation of the ordered target pair;

    \item evaluates the chosen HT or LT maneuver depending on the case study;

    \item returns the transfer duration, propellant consumption, and feasibility indicator; and

    \item removes the action if the trajectory is infeasible or if the available propellant is insufficient.
\end{enumerate}

For an admissible travel action, the fully observed transition is

\begin{equation}
\ell_{e+1}=j,
\qquad
F_{e+1}
=
F_e-\mathcal{D}(m_p),
\label{eq:fuel_transition}
\end{equation}

where $\mathcal{D}$ maps continuous propellant expenditure to the
chosen fuel discretization. A conservative implementation uses

\begin{equation}
\mathcal{D}(m_p)
=
\left\lceil
\frac{m_p}{\Delta F}
\right\rceil.
\label{eq:fuel_discretization}
\end{equation}

The transfer duration is used in the degradation probability in
Eq.~\eqref{eq:degradation} and in the downtime component of the reward
in Eq.~\eqref{eq:reward}. The propellant consumption determines both
the travel cost and the next fuel state.

\section{Solution Algorithm}
\label{sec:solution}

The proposed discrete POMDP is solved offline using the Successive
Approximations of the Reachable Space under Optimal Policies
(SARSOP) algorithm \cite{kurniawati2008sarsop}. The problem is
represented as a mixed-observability Markov decision process (MOMDP),
since the servicing-vehicle location and remaining propellant are
fully observable, whereas the satellite health states are partially
observable. This reduces the belief space to the uncertain health
component of the state \cite{ong2009pomdps}. A detailed description
of the MOMDP formulation and SARSOP solution procedure is provided
in our previous work \cite{patnala2026optimal}.

The main modification in the present work concerns the construction
of the orbital transfers and reward models. For every fully observable
servicer state $x=(\ell,F)$  and travel action
$\mathrm{Travel}_{i\rightarrow j}^{f}$, the current servicer mass is
first calculated using the Eq. \eqref{eq:servicer_mass}.
The corresponding propulsion model is then evaluated as
\begin{equation}
\left(i,j,f,F_e\right)
\xrightarrow{\mathrm{HT\ or\ LT}}
\left(
\Delta t_{ij}^{f},
m_{p,ij}^{f},
\chi_{ij}^{f}
\right).
\label{eq:trajectory_mapping}
\end{equation}

where $i$ and $j$ are the origin and destination satellites, the architecture, HT or LT, determines which transfer table is used. For either architecture, the
policy selects the destination $j$ and travel mode $f$. $F_e$ is the current propellant state, $\Delta t$ is the transfer duration, $m_p$ is the propellant
consumption, and $\chi$ is a maneuver-feasibility indicator. 

These outputs are used to construct the state-dependent transition and reward entries
required by SARSOP. Furthermore, SARSOP accepts a single global discount factor, whereas the present model requires the continuation factor to depend on the duration of the
selected action. To preserve the semi-Markov factor
$\gamma^{\Delta t_a}$, the exported POMDPX uses
$\bar{\gamma}
=
\gamma^{\Delta t_{\min}},
$

where $\Delta t_{\min}$ is the shortest admissible event duration. For
an action of duration $\Delta t_a$, the physical next state is assigned
the continuation probability

\begin{equation}
p_{\mathrm{cont}}(x,a)
=
\frac{\gamma^{\Delta t_a}}{\bar{\gamma}},
\end{equation}

and the remaining probability is assigned to a zero-value terminal
state. Thus, the effective continuation factor used by SARSOP is $
\bar{\gamma}\,p_{\mathrm{cont}}(x,a)
=
\gamma^{\Delta t_a}
$.

This provides a SARSOP-compatible representation of the
duration-dependent discount without changing the solver.

For each propulsion configuration, HT and LT, the discrete states, actions, observations, and transition
probabilities, observation probabilities, and rewards are exported in
the \texttt{.pomdpx} format and solved using SARSOP.

The resulting policy $\pi^\star$ is evaluated through Monte Carlo rollouts in the original POMDP simulator. For each rollout, the true initial health state is sampled from the initial belief, and the true state and belief are propagated according to the selected actions, health transitions, and observations. The evaluation records the discounted return, propellant consumption, failed-satellite time, inspection and servicing actions, transfer durations, and HT and LT utilization.

\subsection{GEO servicing scenario}\label{case_study}

The numerical study considers one servicing vehicle and three target satellites on a common circular GEO orbit. The servicing vehicle is initially co-located with satellite 1 and carries its maximum available propellant. 

The same health model, initial belief, costs, random seeds, and
non-travel action parameters are used in both cases. Differences in performance can therefore be attributed to the propulsion-dependent travel alternatives.

\subsection{Model parameters}
\label{subsec:model_parameters}

Table~\ref{tab:parameters} summarizes the parameters used for the
three-satellite GEO comparison. Parameters from the preceding HT study
were retained where applicable. The servicing vehicle has an initial wet
mass of $700~\mathrm{kg}$, and the travel actions are defined using the HT and LT transfer models described in Section~\ref{sec:transfer}.

\begin{table*}[t]
\centering
\caption{Parameters for the three-satellite GEO comparison.}
\label{tab:parameters}
\small
\renewcommand{\arraystretch}{1.08}
\begin{tabular}{lll}
\toprule
Group & Parameter & Value \\
\midrule
Orbit
& GEO radius; Initial phase angle
& $42164$ km; $0^\circ$, $120^\circ$, $240^\circ$ \\
Servicer
& Dry mass; payload; initial propellant
& $500$ kg; $100$ kg; $100$ kg \\
Fuel state
& Fuel-bin size; number of bins
& $0.25$ kg; $400$ \\
HT
& Specific impulse; computed transfer-duration range
& $400$ s; $1.66$--$6.32$ days \cite{patnala2026optimal} \\
LT
& Thrust; specific impulse; travel durations
& $0.237$ N \cite{herman2007next}; $4190$ s \cite{herman2007next}; $30$, $14$, $10$ days  \\
Events
& Idle; inspection; maintenance duration
& $1$ day; $1$ day; $7$ days \\
Health
& Degradation rate; remote inspection accuracy
& $1/50$ day$^{-1}$; $0.90$ \\
Reward
& Repair reward; maintenance cost; inspection cost
& $20$; $3$; $0.2$ (all in  M\$) \\
Reward
& Fuel; depreciation; downtime cost
& $5/7$ M\$ kg$^{-1}$; $0.05$ M\$ day$^{-1}$; $2$ M\$ per day per satellite\\
POMDP
& discount factor per day; visible states; hidden states
& $0.95$; $1203$; $8$ \\
Actions
& HT; LT (at any given location)
& $11$; $11$ \\
SARSOP
& Precision target; time limit
& $0.1$; $5400$ s \\
Evaluation
& Matched-seed rollouts; horizon
& $1000$ per policy; $100$ events \\
\bottomrule
\end{tabular}
\end{table*}

The policy comparison is conditional on three important modeling
assumptions. First, the daily discount factor is set to
$\gamma=0.95$, corresponding to an effective discounted horizon of
approximately $1/(1-\gamma)=20$ days. This places substantial weight
on the immediate consequences of long-duration actions. Second, the
degradation rate is set to $\lambda_{\mathrm{deg}}=1/50$~day$^{-1}$,
corresponding to a mean time to failure of 50 days under the exponential
model. Third, the available LT transfer durations are restricted to 10, 14, and 30 days. For the initial mass and phase separation considered here, the feasibility condition in Eq.~\eqref{eq:lt_feasibility} permits an LT duration of approximately 7 days; however, transfer alternatives shorter than 10 days are not included in the present action grid. The results thus compare the policies derived under the given parameters and action sets, rather than providing a general ranking of HT versus LT propulsion.

The observable state contains the servicer location and the remaining fuel bucket. There are three possible locations and 401 fuel levels, including
zero fuel, giving $3\times401=1203$ visible states. Each of the three target satellites has a binary healthy/failed condition, producing $2^3=8$ hidden health states. Thus, the exported model contains $1203\times8=9624$ joint state configurations, while maintaining a continuous belief distribution over the eight hidden health states.

The HT model contains three travel modes for each satellite pair. Its
eleven actions comprise idle, local maintenance, three inspection actions, and six travel actions. This is given relative to the current location. The LT model contains the same number of actions. The low, med, and high HT model settings range from approximately 1.66 to 6.32 days. The low, medium, and high LT labels denote the
30-, 14-, and 10-day duration alternatives, respectively. All LT alternatives use the same thrust and specific impulse. 

\section{Results and Discussion}
\label{sec:results}

All policy comparisons reported below are conditional on the discount
factor, degradation rate, and transfer-action grids defined in
Section~\ref{subsec:model_parameters}. The fixed-duration LT transfer implementation was first verified against the eight-day, $180^\circ$ GEO phasing case reported in
Ref.~\cite{sarton2022orbit}. The implementation reproduces the
reported mass-feasibility limit and propellant consumption for this
reference case. The 10-, 14-, and 30-day results are predictions
obtained from the verified analytical model. Appendix \ref{app:lt_validation} summarizes the
verification and these duration-dependent model predictions. The remainder of this section compares the policies obtained using HT and LT propulsion modes. 

The initial wet mass is 700~kg, and the initial propellant considered is 100~kg, with a discretization of 0.25~kg per state bin.  The discounting factor considered is $\gamma=0.95$ with semi-Markov event discounting.  Inspection and maintenance require 1 and 7 days, respectively.  

At an initial wet mass of 700~kg, the HT transfer options require approximately 19-70~kg of propellant and take about 1.7-6.3 days to complete, depending on the selected mode. In contrast, the LT options require only about 0.4-1.3~kg of propellant, but with substantially longer transfer durations of 10-30 days. The comparison, therefore, highlights the expected propulsion trade-off: HT enables significantly faster transfers at the expense of higher propellant consumption, whereas LT greatly reduces propellant expenditure at the cost of longer transfer times.

\subsection{Solver convergence and performance metrics}
\label{subsec:solver_results}

The HT and LT POMDPs were solved independently with SARSOP. Each run used
a requested bound-gap precision of 0.1 and a 5400~s time limit. Both runs
terminated at the time limit. The final bound gaps were 0.107 for HT and
0.140 for LT, with solver wall times of 90.43 and 90.12 minutes,
respectively. The resulting policies are therefore time-limited policy
solutions rather than policies certified to the requested precision.

Each policy was evaluated in 1000 matched-seed Monte Carlo rollouts, with
at most 100 decision events per rollout.  Table~\ref{tab:policy_results} summarizes the solver convergence and
rollout performance. Monte Carlo estimates are presented as sample averages, along with 95\% confidence intervals. The actual propellant consumption, obtained directly from the transfer model, is reported before discretization. The 100-event horizon does not represent the same physical duration for the two policies.  

A few performance metrics are defined and analyzed across both propulsion modes to better understand the policies. Operational availability is defined as the fraction of the required operating time during which a repairable system is capable
of performing its intended function \cite{nasa2017reliability}.
Availability and outage-based measures have also been applied at the
satellite-constellation level \cite{zheng2009availability,
teo2021constellation}.

For each rollout $r$, we define the failed-satellite exposure, equivalently
the cumulative satellite downtime, as

\begin{equation}
D_{\mathrm{fail}}^{(r)}
=
\sum_e \sum_{i=1}^{N_{\mathrm{sat}}}
\int_{0}^{\Delta t_e}
\mathbf{1}\!\left\{h_i^{e}(t)=1\right\}\,dt ,
\label{eq:failed_exposure}
\end{equation}

where $\mathbf{1}\{h_i^{e}(t)=1\}$ is one when satellite $i$ is failed
during event $e$. The quantity, therefore, has units of satellite-days.
A satellite that remains failed throughout a seven-day event contributes
seven satellite-days, while two failed satellites over the same interval
contribute fourteen satellite-days.

The corresponding constellation availability fraction for rollout $r$ is

\begin{equation}
A^{(r)}
=
1-
\frac{D_{\mathrm{fail}}^{(r)}}
     {N_{\mathrm{sat}}T^{(r)}} ,
\label{eq:availability_fraction}
\end{equation}

where $T^{(r)}$ is the total elapsed mission time and
$N_{\mathrm{sat}}T^{(r)}$ is the total available satellite-time.
The reported availability is the arithmetic mean of $A^{(r)}$ over all Monte Carlo rollouts.

\begin{table*}[!t]
\centering
\caption{SARSOP convergence and matched-seed rollout statistics for the
HT and LT policies. Rollout statistics use 1000 simulations, each with at most 100 decision events.}
\label{tab:policy_results}
\small
\renewcommand{\arraystretch}{1.06}
\begin{tabular}{lcc}
\toprule
Metric & HT & LT \\
\midrule
\multicolumn{3}{l}{\textit{SARSOP solution}}\\
Solver termination & Time limit & Time limit \\
Solver wall time [min] & 90.43 & 90.12 \\
Lower bound at $b_0$ & $2.283$ & $1.300$ \\
Upper bound at $b_0$ & $2.390$ & $1.440$ \\
Final bound gap & $0.107$ & $0.14$ \\
\addlinespace
\multicolumn{3}{l}{\textit{Monte Carlo evaluation per rollout}}\\
Discounted rollout return & $1.896\pm1.423$ & $1.232\pm1.064$ \\
Propellant consumed (kg) & 73.5 $\pm$ 1.14 & 5.75 $\pm$ 0.12 \\
Elapsed mission time (days) & 135.83 $\pm$ 0.67 & 192.84 $\pm$ 4.5 \\
Satellite availability fraction & 0.687 $\pm$ 0.009 & 0.619 $\pm$ 0.011 \\
Successful repairs  & 4.34 $\pm$ 0.09 & 5.2 $\pm$ 0.11 \\
Servicer transfers  & 2.78 $\pm$ 0.05 & 4.92 $\pm$ 0.17 \\
Inspections & 44.24 & 59.33 \\
Maintenance actions  & 4.35 & 5.2 \\
Fuel-exhausted rollouts & 116 & 0 \\
\bottomrule
\end{tabular}
\end{table*}

\subsection{Policy-level propulsion trade-off}
\label{subsec:policy_tradeoff}

The computed HT policy achieved the higher finite-horizon discounted
return. Its mean return was $1.896$, compared with $1.232$ for LT, a
difference of $0.664$. The corresponding marginal 95\% confidence
intervals are $[0.473,3.319]$ and $[0.168,2.296]$, respectively. Because the rollouts used matched random seeds, the return comparison was
performed using the paired difference
$D_r=R_{\mathrm{HT},r}-R_{\mathrm{LT},r}$. Across the 1000 paired
rollouts, the mean difference was $0.664$, with a 95\% confidence interval
of $[-0.234,\,1.562]$. 
The estimated average paired return difference leaned in favor of HT, but the 95\%
confidence interval included zero. As a result, the current evaluation does not demonstrate a statistically significant difference in mean return between the HT and LT policies.

LT provided the expected propellant advantage. It consumed
$5.75\pm0.12$~kg per rollout, compared with
$73.50\pm1.14$~kg for HT, corresponding to an approximately $92.2\%$
reduction relative to HT. No LT rollout exhausted the initial propellant supply, whereas 116 HT rollouts did. The HT policy maintained the higher availability fraction, $0.687\pm0.009$, compared with $0.619\pm0.011$ for LT. Thus, LT preserved propellant at the cost of longer transfers and lower constellation availability. However, the mass dependence has only a small numerical effect in the present case: the LT policy consumes approximately \(5.75~\mathrm{kg}\) from an initial \(700~\mathrm{kg}\) spacecraft, and the mass-feasibility constraint is not
binding for any of the available LT transfer alternatives. 

\begin{figure*}[!t]
\centering
\includegraphics[width=\textwidth]{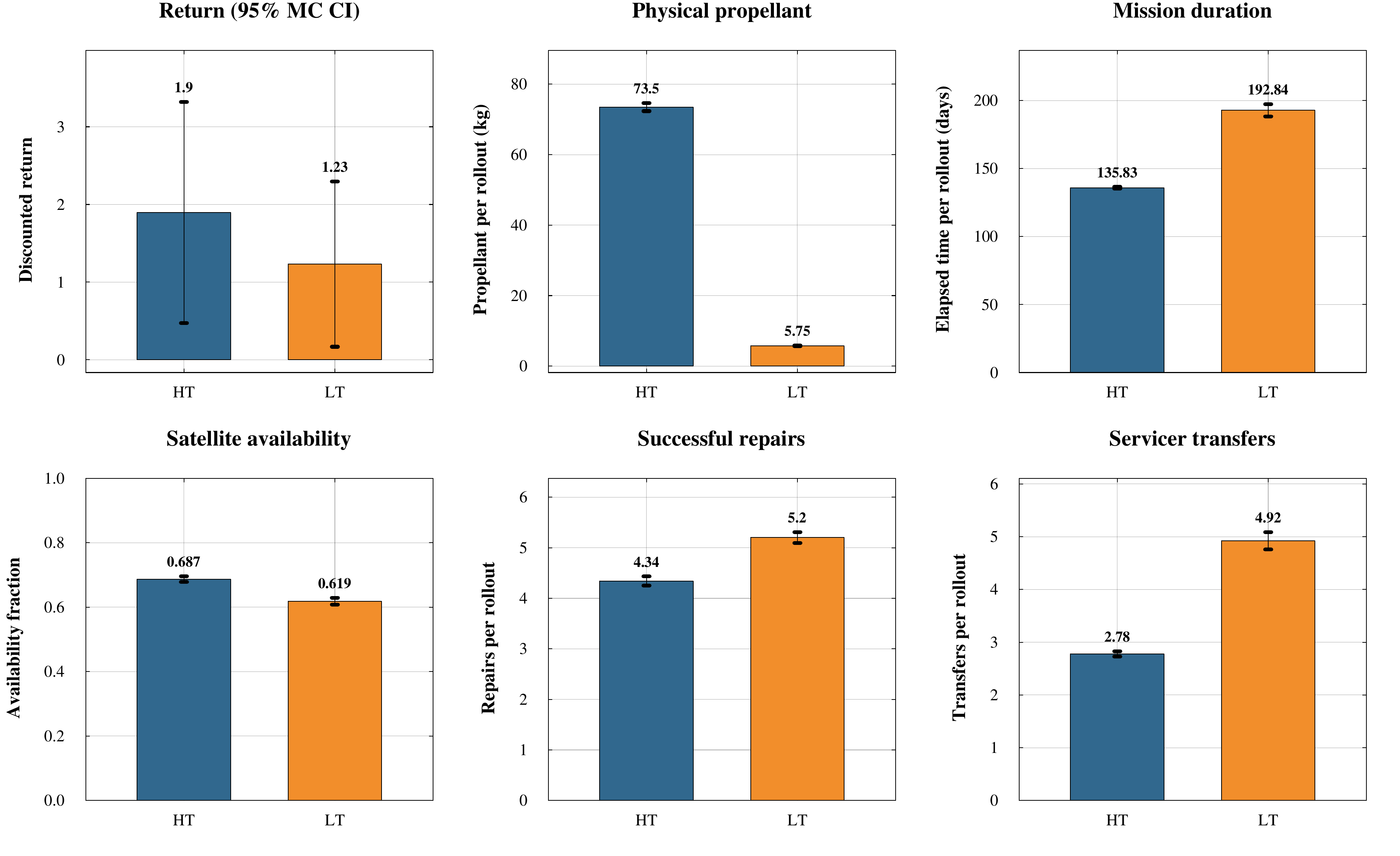}
\caption{Aggregate performance of the HT and LT policies.  Error
bars denote 95\% confidence intervals over 1000 matched-seed rollouts.}
\label{fig:policy_performance}
\end{figure*}

Within the 100-event rollout horizon, the LT policy completed an average
of $5.20$ successful repairs and performed $4.92$ transfers per rollout,
whereas the HT policy completed $4.34$ repairs and performed $2.78$
transfers. The LT rollouts covered an average of $192.84$ days, compared
with $135.83$ days for HT. The longer LT transfer events allow more
degradation and failed-satellite exposure to accumulate before the next
decision epoch, and apply the duration-dependent continuation factor
$\gamma^{\Delta t}$. Although the LT policy completed more repairs over the 100-event rollout, its longer transfer durations delayed maintenance and caused satellites to remain failed for a larger fraction of the mission. Consequently, LT achieved lower availability. HT achieved a higher observed mean return, but the difference is not statistically significant. Figure~\ref{fig:policy_performance} summarizes these performance metrics. 

The satellites accumulated an average failed-satellite exposure of
$130.31$~satellite-days under HT and
$249.76$~satellite-days under LT. This difference is partly
explained by the longer mission duration of the LT rollouts. However, even
after accounting for mission duration, the LT policy results in lower
satellite availability. Thus, although LT transfers consume less propellant,
their longer durations increase satellite degradation and delay maintenance,
which can offset the benefits of improved propellant efficiency.

\subsection{Travel-mode selection and representative behavior}
\label{subsec:mode_selection}
The two policies used different parts of their available action grids.
Across the 1000 rollouts, HT selected the medium mode an average of
approximately $1.95$ times per rollout, and the low mode approximately
$0.83$ times per rollout; the high mode was not selected. LT selected the
10-day high mode approximately $4.06$ times per rollout, and the 30-day
low mode approximately $0.87$ times per rollout; the 14-day medium mode
was not selected. Thus, the HT policy favored the intermediate-duration
transfer, whereas LT primarily selected the fastest LT alternative and
occasionally used the fuel-efficient 30-day transfer. 

\begin{figure*}[!t]
\centering
\includegraphics[width=\textwidth]{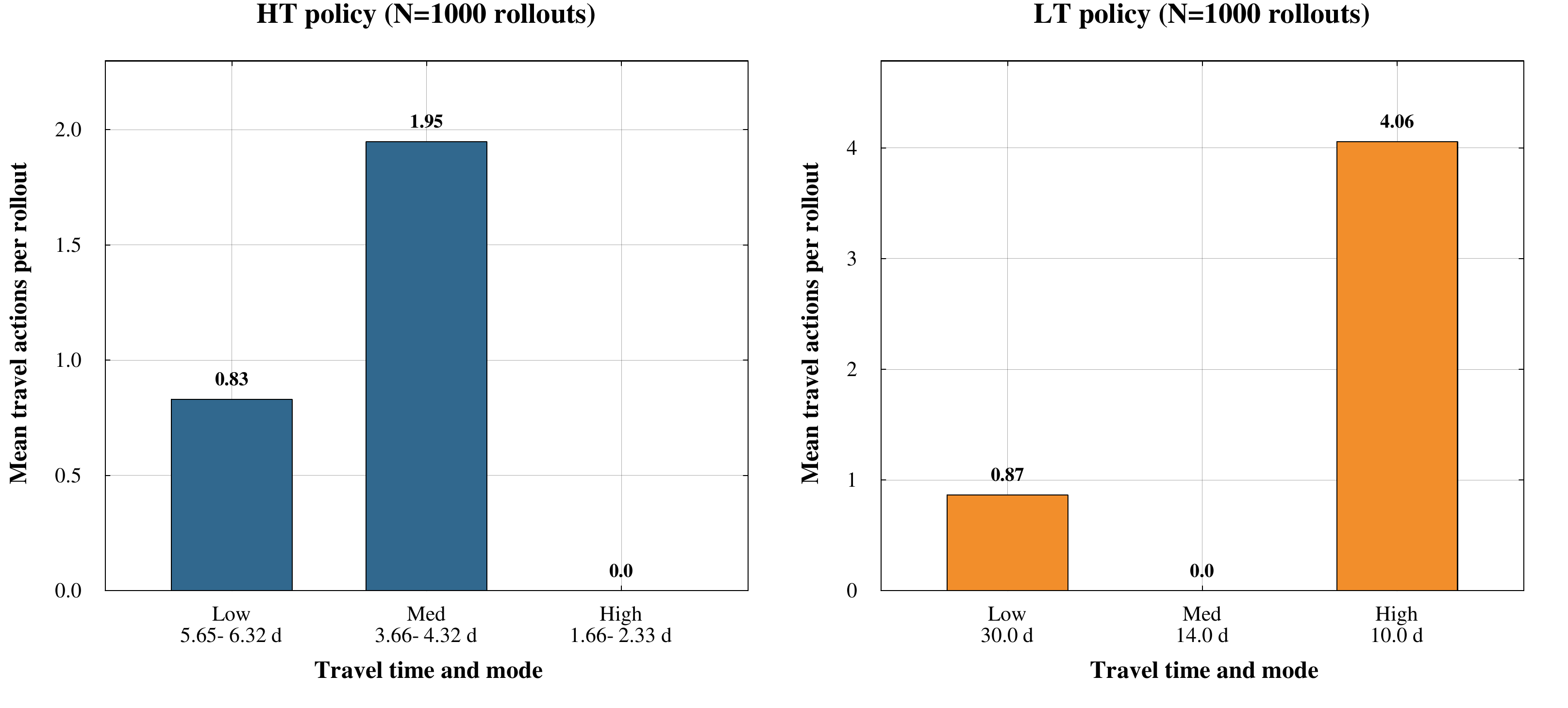}
\caption{Mean travel-mode selections per rollout. Summary of 1000 rollouts. For HT, each mode represents a set of pair-dependent phasing transfers; the duration shown below each mode is the minimum--maximum transfer time across the ordered satellite pairs.  For LT, the modes
correspond to the prescribed fixed durations of 30, 14, and 10 days.}
\label{fig:travel_mode_usage}
\end{figure*}
\begin{figure*}[h!]
\centering
\includegraphics[width=\textwidth]{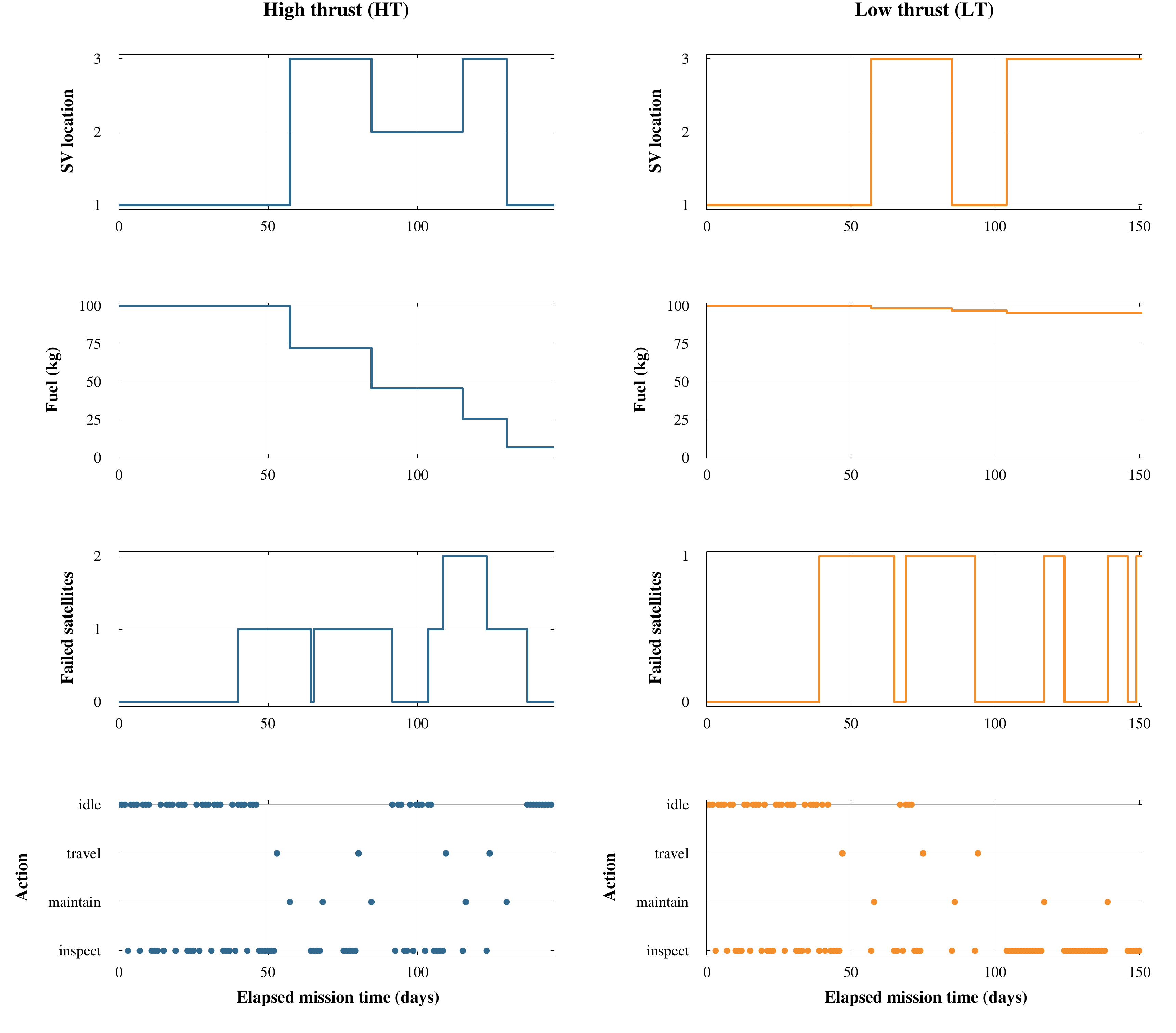}
\caption{Representative matched-seed HT and LT rollouts showing
servicer (SV) location, remaining discretized propellant, realized failed
satellites, and selected actions versus elapsed mission time.}
\label{fig:representative_rollout}
\end{figure*}
Almost every maintenance action resulted in a successful repair. The HT
policy performed $4.35$ maintenance actions and completed $4.34$
successful repairs per rollout, while LT performed $5.20$ maintenance
actions and completed $5.20$ repairs. The policies also performed
$44.24$ inspections under HT and $59.33$ under LT.  The higher LT inspection count is consistent with its longer elapsed mission time and more frequent travel events. During a long LT transfer, the
belief is propagated without receiving a new health observation, and the probability of additional failures increases with the transfer duration. Consequently, propulsion affects not only propellant consumption and downtime, but also the rate at which previously acquired health information becomes outdated. These results indicate that the policies generally gathered information before performing maintenance, and a tiny portion of maintenance actions were performed on healthy satellites (0.01 in HT and 0.001 in LT). Because health is
partially observable, maintenance is selected using the current belief
that the co-located satellite is failed; the true health state is not
directly available to the policy. Consequently, a small number of
maintenance actions may be applied to a healthy satellite. During a long
LT transfer, the belief is propagated without a new health observation,
while the probability of additional failures increases with transfer
duration. This explains an important coupling between the transfer model and the partially observable servicing problem. 

Figure~\ref{fig:representative_rollout} compares one matched-seed realization.  The elapsed-time axes differ because the same number of decision events can span different physical durations.  A common random seed reduces the variance of the aggregate paired comparison, but it does not ensure identical health degradations when the policies choose actions of different durations.  The representative trajectory, therefore, illustrates behavior only; all quantitative conclusions above are based on the complete set of 1000 rollouts.

\section{Conclusions}
\label{sec:conclusion}

This study extends an event-based POMDP for GEO satellite servicing by incorporating propulsion-dependent transfer duration, propellant consumption, and mass feasibility. The semi-Markov formulation accounts for action duration in satellite degradation, cost accumulation, and continuation discounting. The low-thrust implementation was verified against the published eight-day GEO phasing case and subsequently incorporated into the POMDP formulation.

For the three-satellite setting and transfer-action grids considered here, LT reduced mean propellant consumption by approximately 92\%, and none of the 1000 LT rollouts exhausted the initial propellant within the 100-event evaluation horizon. However, its longer transfer durations increased satellite downtime and reduced availability. Consequently, HT had a higher observed mean return, but the difference was not statistically significant. LT retained a substantially larger propellant margin than HT. Although each policy used only two of the three available modes, the selected actions reflect the embedded time--propellant trade-off.

These findings reflect the chosen assumptions regarding discounting, degradation, and transfer and do not provide a definitive overall ranking of HT and LT propulsion. A lower daily discount rate, corresponding to $\gamma$ closer to one, or the inclusion of faster LT alternatives, could change the relative performance of the two policies. Relative performance is determined by how much weight is placed on fast servicing, satellite availability, and saving propellant. Accordingly, the propulsion architecture, feasible transfer options, and reward parameters should be chosen in an integrated manner. Under partial observability, both policies used inspections to update their beliefs about satellite health. The LT policy conducted more inspections than the HT policy, consistent with the longer decision-epoch intervals and the greater time elapsed since the most recent health observations.

Future work will quantify the sensitivity of the results to the discount
factor, degradation rate, reward parameters, and transfer-action grid.
It will also consider fully observable and rule-based baselines, improved
inspection models, higher-fidelity trajectory models, and larger servicing scenarios.
\section{Acknowledgments}
This work is supported by the \emph{Agence Nationale de la Recherche (ANR)}, France, under the project \textbf{ANR-23-CE10-0006} (Resilient and Sustainable Planning and Management of Future Space Industry Infrastructure with On-Orbit Servicing), for which Prof. Adam Abdin is the principal investigator. Travel expenses partially covered by a travel award sponsored by the open-access journal \emph{Aerospace} published by MDPI.
\balance
\printbibliography
\appendix
\section{Verification of the low-thrust transfer model}
\label{app:lt_validation}

The fixed-duration LT implementation was verified separately from
the POMDP using the eight-day, $180^\circ$ GEO phasing case reported
in Ref.~\cite{sarton2022orbit}. The reference case uses

\begin{equation}
F_T=1.16~\mathrm{N},\qquad
I_{sp,\mathrm{LT}}=1790~\mathrm{s},\qquad
\Delta t=8~\mathrm{days},
\end{equation}

with $r_0=42164~\mathrm{km}$ and $500\leq m_0\leq4000~\mathrm{kg}$.
The implementation gives a maximum feasible initial mass of
$3137.9~\mathrm{kg}$, consistent with the reported value of
$3138~\mathrm{kg}$. At $m_0=1800~\mathrm{kg}$, the predicted propellant
consumption is $15.8512~\mathrm{kg}$.

\begin{figure}[!t]
\centering
\includegraphics[width=\columnwidth]{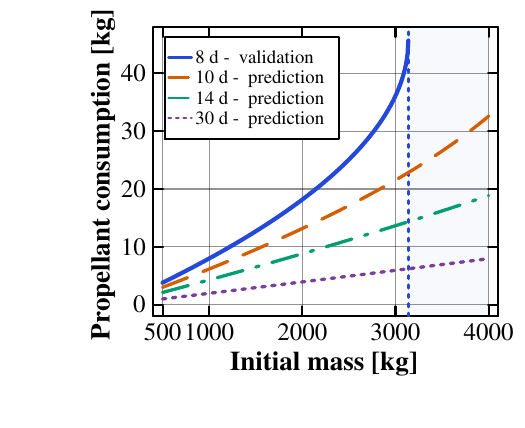}
\caption{Fixed-duration LT propellant curves evaluated for the
reference $180^\circ$ GEO configuration, with
$F_T=1.16$~N and $I_{sp,\mathrm{LT}}=1790$~s. The eight-day curve reproduces the published case in
Ref.~\cite{sarton2022orbit}.}
\label{fig:lt_reference_validation}
\end{figure}

Figure~\ref{fig:lt_reference_validation} also shows predictions for
durations of 10, 14, and 30 days. These are the alternate travel durations used in the POMDP study, but all curves in Fig.~\ref{fig:lt_reference_validation} are evaluated using the reference
thruster parameters, $F_T=1.16$~N and $I_{sp,\mathrm{LT}}=1790$~s \cite{sarton2022orbit}.
At $m_0=1800~\mathrm{kg}$, these require $11.6741$, $7.8740$, and
$3.5299~\mathrm{kg}$, respectively. Only the eight-day curve is a direct
reproduction of the published validation case; the remaining curves are
predictions from the verified model using the given LT parameters. 

\end{document}